\documentclass[11pt]{article}
\usepackage[preprint]{acl}
\usepackage{times}
\usepackage{latexsym}
\usepackage[T1]{fontenc}
\usepackage[utf8]{inputenc}
\usepackage{microtype}
\usepackage{booktabs,amsmath,amssymb,graphicx,multirow}
\usepackage{tikz}
\usetikzlibrary{arrows.meta,positioning}
\newcommand{\graphecho}{GraphEcho}
\newcommand{\method}{PAPT}
\title{\graphecho{}: Structural Redundancy and Evidence Provenance \\in LLM Graph Agents}

\author{
\textbf{Sikun Wang}\textsuperscript{1}\thanks{Equal contribution} \quad
\textbf{Yixi Zhou}\textsuperscript{2}\footnotemark[1] \quad
\textbf{Lei Fan}\textsuperscript{3} \quad
\textbf{Fan Zhang}\textsuperscript{4}\thanks{Corresponding author} \\
\textsuperscript{1}Tokyo University of Science \quad
\textsuperscript{2}Hong Kong Baptist University \\
\textsuperscript{3}University of Michigan \quad
\textsuperscript{4}The University of Tokyo \\
yxzhou@comp.hkbu.edu.hk, jc19546883@gmail.com \\
leifanus@gmail.com, zhang-fan@g.ecc.u-tokyo.ac.jp
}

\begin{document}
\maketitle
\begin{abstract}
A large language model (LLM) agent can follow more graph paths without acquiring more independent evidence. \graphecho{} tests whether agents mistake these repeated encounters for additional corroboration. The benchmark varies path counts and evidential origins while holding evidence content fixed, and evaluates both judgments and active exploration. Controlled synthetic experiments reveal model-dependent judgment shifts, but redundant supporting paths increase the share of repeated walks across all evaluated frozen agents. Provenance-aware post-training (PAPT) reduces revisits and improves synthetic accuracy, yet covers fewer distinct sources. On scientific claims, it continues to reduce repetition while accuracy declines. These findings expose a gap between efficient exploration and effective evidence use: an agent can learn to stop repeating itself while overlooking information it needs. \graphecho{} provides a controlled way to evaluate both what graph agents conclude and whether their exploration reaches distinct evidential sources.
\end{abstract}
\section{One observation can yield many paths}
A graph agent can encounter one experimental observation through a result node, a conclusion node, and a table entry. These paths share one evidential origin. Counting them as independent corroboration would change the agent's assessment without adding evidence. Repeated retrieval can also consume its exploration budget.

Retrieval-augmented reasoning connects evidence selection with intermediate decisions~\citep{trivedi2023ircot,asai2024selfrag}, while graph retrieval links information across passages~\citep{gutierrez2024hipporag}. Recent work examines repeated retrieval contexts~\citep{ross2026redundancy}, learned graph exploration~\citep{liu2026cny}, and provenance-aware evidence selection~\citep{deng2026page}. These directions motivate a controlled test of how structural representations affect both judgment and evidence acquisition.

\begin{figure*}[t]
\centering
\includegraphics[width=\textwidth]{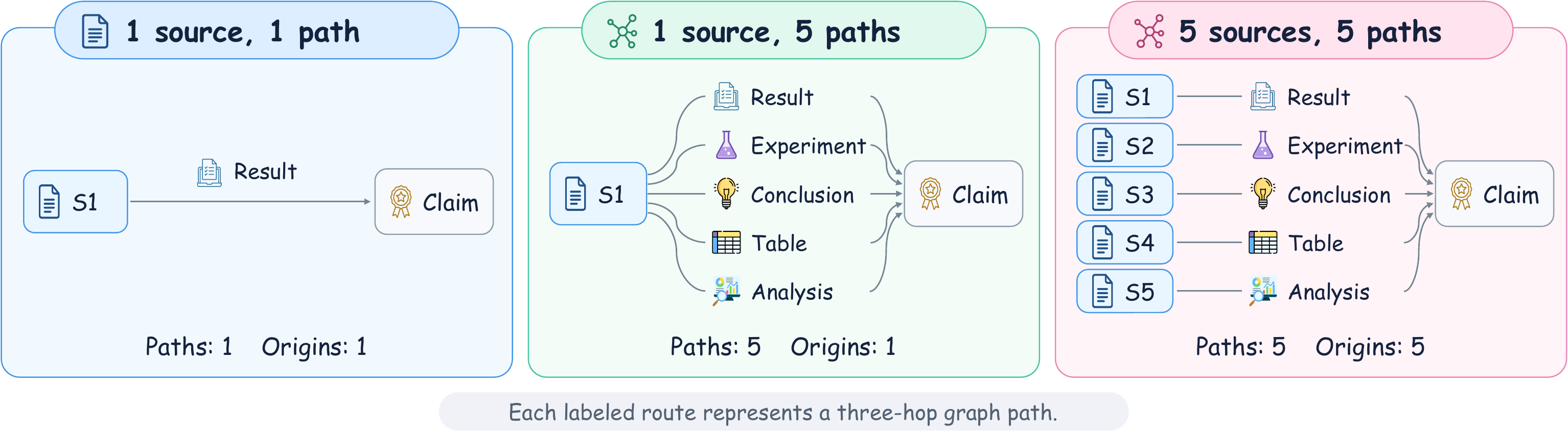}
\caption{Path count and provenance count define separate interventions. Each route contracts a three-hop path to the same claim; the five-route comparison preserves representation types and evidence wording.}
\label{fig:control}
\end{figure*}

\graphecho{} measures whether an agent distinguishes structural multiplicity from provenance multiplicity. Our central comparison holds the claim and five path templates fixed while changing whether the paths originate from one study or five separate studies. A second comparison expands one path into five paths from the same study. Together, these interventions separate sensitivity to additional representations from sensitivity to additional evidential origins (Figure~\ref{fig:control}).

This comparison requires control over evidence content, source strength, and graph presentation. Different path lengths could change reasoning difficulty, while different evidence statements could add information. We therefore match path lengths and evidence wording in the synthetic source-count intervention. 


Our contributions establish three findings:
\begin{itemize}
\item \textbf{Controlled judgment.} \graphecho{}-Syn separates redundant paths from distinct origins across 800 fictional claims; static responses differ in direction across models.
\item \textbf{Active acquisition.} Redundant supporting paths raise echo walk rates in all four frozen agents, while accuracy changes in both directions.
\item \textbf{Training tradeoffs.} Provenance-aware post-training (\method) raises synthetic agent accuracy by 41.5 percentage points over frozen Qwen3-4B, but lowers source recall and SciFact accuracy. Cheap controls and reward ablations distinguish reduced repetition from improved evidence use.
\end{itemize}

\section{Related Work}
\paragraph{Evidence conflict and redundancy.}
Conflicting passages can change answers without a corresponding reduction in model confidence~\citep{chen2022conflicts}. \citet{tan2024blinded} find that models favor generated over retrieved contexts under knowledge conflict, while \citet{yoran2024robust} improve tolerance to irrelevant passages through training. Prior work studies context position~\citep{liu2024lost} and retrieval redundancy~\citep{ross2026redundancy}. \graphecho{} isolates a different source of distortion: multiple graph paths can repeat one evidential origin. Matching evidence wording and path templates separates structural repetition from source count.

\paragraph{Attribution and graph retrieval.}
ALCE evaluates answer correctness and citation quality separately~\citep{gao2023alce}. HippoRAG combines knowledge graphs with Personalized PageRank for multi-hop retrieval~\citep{gutierrez2024hipporag}; HippoRAG~2 extends this approach with passage integration~\citep{gutierrez2025memory}. PAGE-RAG tracks provenance when selecting graph evidence~\citep{deng2026page}. These methods motivate testing whether accessible source information suffices to prevent duplicate corroboration. \graphecho{} varies provenance assignments while holding evidence content fixed, and compares visible source identifiers with preprocessing that collapses repeated evidence.

\paragraph{Adaptive evidence acquisition.}
ReAct interleaves reasoning with actions~\citep{yao2023react}, and IRCoT alternates retrieval with intermediate reasoning steps~\citep{trivedi2023ircot}. Adaptive-RAG selects retrieval strategies according to question complexity~\citep{jeong2024adaptive}, while Self-RAG learns when to retrieve and how to assess retrieved passages and generated answers~\citep{asai2024selfrag}. Call Neighbours Yourself (CNY) trains graph-neighbor selection through destination-conditioned on-policy self-distillation~\citep{liu2026cny}. \graphecho{} examines whether acquisition reaches distinct origins within a fixed action budget. PAPT uses group-relative policy optimization~\citep{shao2024deepseekmath} with rewards for judgments, source counting, and distinct evidence acquisition. Evaluating accuracy together with repeated walks and source recall distinguishes fewer revisits from broader evidence coverage.

\section{Separate paths from evidential origins}
\label{sec:benchmark}
\subsection{Provenance defines the intervention}
An evidence provenance atom $e$ identifies an underlying evidential origin. For a graph path $p$, the mapping $\pi(p)=e$ records that origin. Given a set of paths $P$, structural and provenance multiplicity are
\begin{align}
N_{\mathrm{path}}&=|P|,\\
N_{\mathrm{prov}}&=|\{\pi(p):p\in P\}|.
\end{align}
We write $s$S$k$P for a condition with $s$ distinct origins and $k$ paths. Redundant-path interventions change $N_{\mathrm{path}}$ while preserving the evidence atoms and their content. Source-count interventions change $N_{\mathrm{prov}}$ at a fixed path count.

\subsection{Synthetic graphs control provenance}
\graphecho{}-Syn contains 800 claims across biomedicine, materials, agriculture, and economics. Fictional entity names separate claim content from familiar factual associations. Each claim has a candidate pool of five supporting atoms and five refuting atoms. The generator assigns every atom a distinct source identifier and preserves both source and atom identifiers through graph construction.

All atoms have the same relative reliability weight, 1.0. This weight expresses equal strength within the benchmark; it does not make an observation infallible. Supporting atoms report a change in the claim's direction, and refuting atoms report the opposite change. The gold verdict follows the balance of distinct origins selected for the displayed condition: SUPPORT for a supporting majority, REFUTE for a refuting majority, and MIXED for a tie. This label describes the evidence balance under the benchmark convention.

Each atom admits up to five three-hop paths through different representation types. The primary source-count comparison uses identical evidence wording and relation templates for 1S5P and 5S5P. Only the origin assignments change. Across variants of the same claim, the generator also preserves claim wording. Token counts remain an explicit diagnostic because tokenization can treat source identifiers differently.

The split contains 500 training, 100 development, and 200 test claims. Every variant of a base claim stays in that claim's split. Entity names also remain disjoint across splits. The initial 100-claim pilot uses development claims, preserving the test split for the subsequent evaluation.

\begin{table}[t]
\centering\small
\setlength{\tabcolsep}{3pt}
\begin{tabular}{lccc}
\toprule
Condition & Support & Refute & Gold\\
 & origins $\times$ routes & origins $\times$ routes &\\
\midrule
C1 & $1\times1$ & $1\times1$ & MIXED\\
C2 & $1\times5$ & $1\times1$ & MIXED\\
C3 & $5\times1$ & $1\times1$ & SUPPORT\\
C4 & $1\times1$ & $5\times1$ & REFUTE\\
C2R & $1\times1$ & $1\times5$ & MIXED\\
\bottomrule
\end{tabular}
\caption{Conflict interventions vary path counts and provenance counts separately.}
\label{tab:conditions}
\end{table}

\subsection{Conflict exposes false corroboration}
Table~\ref{tab:conditions} specifies five conflict conditions. C1 and C2 contain the same two opposing atoms. C2 adds four representations of the supporting atom, so an increased SUPPORT rate measures a decision shift under redundant evidence. C3 supplies five distinct supporting origins. C4 reverses the provenance majority, and C2R checks whether redundant refuting paths produce a corresponding shift.

\subsection{SciFact tests document provenance}
SciFact pairs scientific claims with annotated evidence abstracts and rationale sentences~\citep{wadden2020scifact}. We use its labeled development split for evaluation because the official test labels are unavailable. The conversion retains 188 claims with gold evidence and produces 584 graph instances. Ten eligible claims have multiple evidence documents. Claims without gold evidence do not enter this gold-rationale experiment.

For each document, the converter preserves the annotated rationale sets and collects their sentences into one evidence record. Redundant routes repeat that record through different graph representations. Document identifiers provide observable source provenance in this experiment. They do not establish statistical independence between scientific studies. \method{} uses synthetic supervision only, leaving SciFact as a transfer evaluation.

\section{Acquire distinct evidence within a budget}
\subsection{Observe graph actions}
The agent environment exposes a current node, visited nodes, available neighbors, short previews, and the remaining walk budget. A walk action reveals the selected neighbor's full content. A stop action returns the final judgment. The primary budget permits six walk actions; additional runs use budgets of three, five, and eight. Each three-hop provenance path connects a source, a representation, an observation, and the claim. Explicit shared-claim edges connect observation nodes, permitting direct movement to another candidate observation. Every action follows an edge; the trace records connector movements separately from repeated evidence acquisition.

The agentic comparison uses balanced provenance, redundant supporting routes, and distinct supporting origins. Neighbor previews withhold unseen evidence content and its gold stance. Provenance metadata may remain visible under the corresponding mitigation condition. An atom's first evidence-bearing visit supplies its complete synthetic observation, making later visits to equivalent representations redundant under this controlled setup.

\subsection{Compare cheap mitigations}
Four inference controls test whether training adds value beyond metadata, instructions, and preprocessing. M0 displays source nodes without atom identifiers. M1 adds provenance identifiers; M2 also instructs the model to count distinct origins when assessing corroboration. M3 groups equivalent paths while preserving their multiplicity.

\subsection{Train a provenance-aware policy}
\method{} combines a verdict reward with source counting and evidence acquisition:
\begin{align}
R={}&R_{\mathrm{verdict}}+0.4R_{\mathrm{count}}\nonumber\\
&+0.4R_{\mathrm{coverage}}-0.4R_{\mathrm{echo}}
-0.05R_{\mathrm{cost}}.
\end{align}
The verdict reward equals $+1$ for a correct final judgment and $-1$ otherwise. The count reward measures agreement with gold supporting and refuting provenance counts. Coverage rewards acquisition of distinct useful origins; the echo penalty measures repeated provenance acquisition. The cost penalty divides the walk count by the budget. With the counts defined in Section~\ref{sec:experiments}, coverage equals $U/\max(1,\min(B,D))$, echo equals $E/\max(1,W)$, and cost equals $W/B$. Let $\Delta$ sum the absolute support and refute count errors over acquired origins. The count reward equals $\max(0,1-\Delta/\max(1,U))$ for a valid final output and zero for an invalid output.

The training protocol applies group-relative policy optimization (GRPO)~\citep{shao2024deepseekmath} to Qwen3-4B with groups of four trajectories. We apply low-rank adaptation (LoRA)~\citep{hu2022lora} with rank 16 and scaling parameter $\alpha=32$. Each training claim contributes one conflict instance, giving 100 examples per condition across C1, C2, C3, C4, and C2R. Training uses these 500 examples for one epoch with random seed 42.

All reward variants share the selected training instances and optimization settings. Ablations remove the echo reward, remove the coverage reward, or retain only the verdict reward. Checkpoint selection maximizes mean full-objective reward on development claims among the final three retained checkpoints. Using the same selection objective across variants makes their development scores comparable. Test labels and SciFact supervision never enter checkpoint selection.

\section{Measure judgment and acquisition separately}
\label{sec:experiments}
\subsection{Models and outputs}
The frozen evaluation covers Qwen3-1.7B, Qwen3-4B, and Qwen3-8B.\footnote{Official model cards: \href{https://huggingface.co/Qwen/Qwen3-1.7B}{Qwen3-1.7B}, \href{https://huggingface.co/Qwen/Qwen3-4B-Instruct-2507}{Qwen3-4B}, and \href{https://huggingface.co/Qwen/Qwen3-8B}{Qwen3-8B}.} The 1.7B and 8B models use non-thinking generation. The primary open-model setting uses greedy decoding. GPT-5-mini\footnote{OpenAI, \href{https://openai.com/index/introducing-gpt-5/}{Introducing GPT-5}.} provides a supplementary API baseline with its native sampling settings, which differ from temperature-zero decoding.

Outputs contain a verdict, confidence, supporting and refuting provenance counts, cited source identifiers, and a short rationale. Confidence means subjective claim probability, not confidence in the selected verdict. Parsing failures count as incorrect verdicts; confidence estimates use valid pairs with reported sample counts.

\subsection{Static effects use paired claims}
Let $C(s\mathrm{S}k\mathrm{P})$ denote confidence for the corresponding graph condition. We measure path multiplicity and independent-source effects as
\begin{align}
\mathrm{PME}_k &= C(1\mathrm{S}k\mathrm{P})-C(1\mathrm{S}1\mathrm{P}),\\
\mathrm{ISE}_k &= C(k\mathrm{S}k\mathrm{P})-C(1\mathrm{S}k\mathrm{P}).
\end{align}
The primary comparison uses $k=5$, with $k=3$ as an additional condition. For one-sided supporting evidence, positive PME indicates increased confidence from redundant paths. ISE measures the response to distinct sources at a fixed path count. We report both effects separately; we report their ratio only when ISE exceeds 0.5 percentage points to avoid an unstable denominator.

Evidence count inflation subtracts the true provenance count from the model's predicted count. Topology flip rate measures verdict changes between 1S1P and 1S5P. Conflict bias compares the SUPPORT rate and claim confidence in C2 against C1. The mirrored C2R condition checks the direction of the corresponding refuting-path effect.

\subsection{Acquisition measures expose repeated visits}
Let $W$ count all walk attempts, $V$ count valid visits to evidence-bearing nodes, and $U$ count distinct acquired origins. Every selected atom bears on the claim and counts as useful, including refuting atoms. Let $D$ count graph-reachable origins, and let $E$ count valid visits whose origin appeared in an earlier visit. We compute
\begin{align}
\mathrm{UPC}&=U/V, & \mathrm{EWR}&=E/W,\\
\mathrm{USR}&=U/D, & \mathrm{WE}&=U/W.
\end{align}
These ratios measure unique provenance coverage (UPC), echo walk rate (EWR), unique source recall (USR), and walk efficiency (WE). They equal zero for zero denominators. Connector and invalid actions enter $W$; $V$ counts only evidence-bearing visits. Appendix~\ref{app:diagnostics} separates connector and invalid-action counts. Shared-claim edges in the primary graph make each origin accessible in one acquisition action, allowing $\min(B,D)$ distinct origins within budget $B$.

Agent source-count error compares predicted counts with the distinct origins acquired. Final verdict accuracy uses the full graph's evidence-balance label. This distinction separates errors about observed provenance from incomplete evidence acquisition.

\subsection{Mitigation must preserve source sensitivity}
A reduction in redundant-path sensitivity is insufficient if a model also stops responding to distinct sources. The mitigation comparison therefore reports PME and ISE together with verdict accuracy and source-count error. Agentic evaluation adds coverage, repeated walks, and cost. SciFact reports SUPPORT and REFUTE strata separately. Its stance-aligned PME multiplies the raw probability change by $+1$ for supporting evidence and $-1$ for refuting evidence. Document-count effects use only eligible multi-document claims and compare matched path counts; we call these document-provenance effects (DPE). These comparisons test transfer without scientific-data supervision.

\subsection{Uncertainty respects repeated measurements}
The main evaluation randomizes path order across five seeds. We average repeated measurements within each base claim before resampling claims. Tables show point estimates with bracketed lower and upper bounds of 95\% confidence intervals, using 10,000 claim-bootstrap resamples. For binary verdict comparisons, the analysis also computes exact McNemar tests using one pair per claim at the lowest presentation seed and sample. The static sampling sensitivity analysis uses temperature 0.7 and five outputs per instance at presentation seed zero. Primary synthetic comparisons use five presentation seeds; comparisons with inference-time controls use their common seed zero. SciFact agent evaluation uses seed zero. Tables state the applicable pairing scope.

Appendix~\ref{app:diagnostics} reports valid-final and zero-walk counts alongside acquisition measures. A post hoc accuracy diagnostic retains only matched condition, seed, and sample pairs with valid finals on both sides. Selection changes which observations enter this comparison and cannot isolate a causal reasoning effect. Primary accuracy includes all received outputs, counting malformed finals as incorrect.

\section{Redundancy changes judgment and acquisition}
\label{sec:results}
\subsection{Static effects depend on the model}
\begin{table*}[t]
\centering\small\setlength{\tabcolsep}{2pt}
\begin{tabular}{lrrrrrrr}
\toprule
Model & 1S1P & 1S5P & 5S5P & PME$_5$ & ISE$_5$ & Count inflation & Flip rate\\
\midrule
Qwen3-1.7B & 90.00 & 90.00 & 90.06 & \shortstack{0.01\\{\footnotesize [0.00, 0.01]}} & \shortstack{0.05\\{\footnotesize [0.01, 0.10]}} & \shortstack{0.01\\{\footnotesize [0.00, 0.02]}} & \shortstack{0.00\\{\footnotesize [0.00, 0.00]}}\\
Qwen3-4B & 87.53 & 95.00 & 95.42 & \shortstack{7.47\\{\footnotesize [7.12, 7.83]}} & \shortstack{0.42\\{\footnotesize [0.29, 0.56]}} & \shortstack{-1.00\\{\footnotesize [-1.00, -1.00]}} & \shortstack{0.00\\{\footnotesize [0.00, 0.00]}}\\
Qwen3-8B & 85.00 & 83.15 & 89.14 & \shortstack{-1.85\\{\footnotesize [-2.29, -1.44]}} & \shortstack{6.00\\{\footnotesize [5.71, 6.29]}} & \shortstack{4.00\\{\footnotesize [4.00, 4.00]}} & \shortstack{0.00\\{\footnotesize [0.00, 0.00]}}\\
GPT-5-mini & 76.16 & 71.67 & 79.63 & \shortstack{-4.49\\{\footnotesize [-5.06, -3.91]}} & \shortstack{7.96\\{\footnotesize [7.46, 8.48]}} & \shortstack{-0.03\\{\footnotesize [-0.04, -0.02]}} & \shortstack{23.60\\{\footnotesize [19.80, 27.40]}}\\
\bottomrule
\end{tabular}
\caption{Static synthetic test evaluation. Probabilities and flip rates use percent; PME and ISE use percentage points. Brackets show 95\% paired base-claim bootstrap intervals. Each model uses 200 test claims and five presentation seeds. Count inflation refers to 1S5P.}
\label{tab:static}
\end{table*}

Static responses differ across models (Table~\ref{tab:static}). Qwen3-4B responds more to redundant paths than to additional origins, whereas Qwen3-1.7B barely changes either probability. Qwen3-8B and GPT-5-mini lower probability under redundant paths and raise it under distinct origins. Source counting reveals another distinction: at 1S5P, Qwen3-8B overcounts by four origins, while Qwen3-4B undercounts by one. The same intervention therefore produces different judgment and counting errors across models.

Conflict conditions separate categorical and probability responses. Qwen3-1.7B shifts strongly toward SUPPORT when redundant supporting paths enter C2. Qwen3-8B and GPT-5-mini keep their SUPPORT rates unchanged, although their claim probabilities rise. Repetition can therefore shift expressed confidence while leaving the categorical decision unchanged. Path sensitivity, source counting, and categorical judgment therefore capture different failures. Appendix~\ref{app:static} reports the complete conflict contrasts and confidence curves.

\subsection{Redundant paths raise echo walk rates}
\begin{table*}[t]
\centering\small\setlength{\tabcolsep}{3pt}
\begin{tabular}{lrrr}
\toprule
Model & $\Delta$ echo rate & $\Delta$ source recall & $\Delta$ accuracy\\
\midrule
Qwen3-1.7B & \shortstack{+6.84\\{\footnotesize [5.62, 8.09]}} & \shortstack{+6.55\\{\footnotesize [4.35, 8.75]}} & \shortstack{+8.80\\{\footnotesize [5.70, 12.00]}}\\
Qwen3-4B & \shortstack{+38.84\\{\footnotesize [37.02, 40.67]}} & \shortstack{-10.95\\{\footnotesize [-12.70, -9.15]}} & \shortstack{-11.00\\{\footnotesize [-15.80, -6.20]}}\\
Qwen3-8B & \shortstack{+19.63\\{\footnotesize [18.59, 20.69]}} & \shortstack{+24.00\\{\footnotesize [21.40, 26.60]}} & \shortstack{+30.30\\{\footnotesize [25.60, 35.00]}}\\
GPT-5-mini & \shortstack{+5.80\\{\footnotesize [4.10, 7.48]}} & \shortstack{-31.20\\{\footnotesize [-33.10, -29.25]}} & \shortstack{-61.90\\{\footnotesize [-65.80, -58.00]}}\\
\bottomrule
\end{tabular}
\caption{Redundant support raises echo walk rates in all four frozen agents. C2-minus-C1 differences use percentage points, 200 claims and five presentation seeds; brackets show 95\% paired claim-bootstrap intervals. Accuracy counts malformed received outputs as incorrect.}
\label{tab:frozenecho}
\end{table*}

All four frozen agents increase their echo walk rate when C2 replaces C1 (Table~\ref{tab:frozenecho}). The paired increases range from 5.80 to 38.84 percentage points, and all four intervals exclude zero. Accuracy changes in both directions: Qwen3-4B and GPT-5-mini lose accuracy, while the other models gain it. Echo walk rate therefore provides a more consistent response to redundant topology than accuracy degradation.

Output validity complicates the accuracy comparison for Qwen3-1.7B. Valid finals rise from 39.1\% in C1 to 90.5\% in C2. Restricting the analysis to pairs with valid finals reverses the estimated accuracy change, but its interval spans zero. This selected subset covers 168 of 200 claims (Appendix~\ref{app:diagnostics}). The primary accuracy gain thus depends on which outputs enter the analysis.

\newpage
\subsection{PAPT learns shorter, less repetitive walks}
\begin{table*}[t]
\centering\small\setlength{\tabcolsep}{2pt}
\begin{tabular}{lrrrrrr}
\toprule
Policy & Accuracy & UPC & Echo rate & Source recall & Efficiency & Count error\\
\midrule
M0: source nodes & \shortstack{43.12\\{\footnotesize [39.62, 46.50]}} & \shortstack{57.36\\{\footnotesize [55.99, 58.69]}} & \shortstack{40.52\\{\footnotesize [39.46, 41.59]}} & \shortstack{90.15\\{\footnotesize [88.77, 91.44]}} & \shortstack{56.03\\{\footnotesize [54.50, 57.50]}} & \shortstack{1.55\\{\footnotesize [1.45, 1.65]}}\\
M1: provenance & \shortstack{42.50\\{\footnotesize [39.12, 45.88]}} & \shortstack{56.44\\{\footnotesize [55.18, 57.68]}} & \shortstack{41.97\\{\footnotesize [40.95, 42.99]}} & \shortstack{87.69\\{\footnotesize [86.29, 89.04]}} & \shortstack{55.47\\{\footnotesize [54.08, 56.81]}} & \shortstack{1.65\\{\footnotesize [1.55, 1.75]}}\\
M2: instruction & \shortstack{42.12\\{\footnotesize [38.75, 45.62]}} & \shortstack{56.61\\{\footnotesize [55.38, 57.83]}} & \shortstack{41.72\\{\footnotesize [40.70, 42.72]}} & \shortstack{86.85\\{\footnotesize [85.46, 88.23]}} & \shortstack{55.45\\{\footnotesize [54.04, 56.81]}} & \shortstack{1.58\\{\footnotesize [1.48, 1.68]}}\\
M3: collapse & \shortstack{83.25\\{\footnotesize [81.00, 85.50]}} & \shortstack{100.00\\{\footnotesize [100.00, 100.00]}} & \shortstack{0.00\\{\footnotesize [0.00, 0.00]}} & \shortstack{95.88\\{\footnotesize [95.75, 95.98]}} & \shortstack{100.00\\{\footnotesize [100.00, 100.00]}} & \shortstack{0.36\\{\footnotesize [0.31, 0.42]}}\\
PAPT & \shortstack{87.12\\{\footnotesize [85.00, 89.25]}} & \shortstack{95.83\\{\footnotesize [94.73, 96.85]}} & \shortstack{4.17\\{\footnotesize [3.15, 5.27]}} & \shortstack{80.48\\{\footnotesize [79.56, 81.31]}} & \shortstack{95.83\\{\footnotesize [94.73, 96.85]}} & \shortstack{0.03\\{\footnotesize [0.01, 0.04]}}\\
\bottomrule
\end{tabular}
\caption{Qwen3-4B agentic evaluation with budget six. All methods use presentation seed zero. We average C1, C2, C3, and C2R within each claim before computing estimates and bootstrap intervals. Rates use percent; count error uses acquired evidence. Brackets show 95\% base-claim bootstrap intervals.}
\label{tab:agentic}
\end{table*}

\begin{table*}[t]
\centering\small\setlength{\tabcolsep}{3pt}
\begin{tabular}{lrrrrr}
\toprule
Full PAPT minus & Syn. accuracy & Syn. echo rate & Syn. source recall & SciFact accuracy & SciFact echo rate\\
\midrule
M1: provenance & \shortstack{+41.47\\{\footnotesize [39.42, 43.50]}} & \shortstack{-36.89\\{\footnotesize [-37.51, -36.28]}} & \shortstack{-8.32\\{\footnotesize [-9.05, -7.59]}} & \shortstack{-3.55\\{\footnotesize [-6.21, -1.06]}} & \shortstack{-36.00\\{\footnotesize [-37.16, -34.80]}}\\
M2: instruction & \shortstack{+45.00\\{\footnotesize [40.88, 49.12]}} & \shortstack{-37.55\\{\footnotesize [-38.86, -36.21]}} & \shortstack{-6.38\\{\footnotesize [-7.90, -4.88]}} & \shortstack{-3.37\\{\footnotesize [-6.21, -0.71]}} & \shortstack{-36.40\\{\footnotesize [-37.43, -35.35]}}\\
M3: collapse & \shortstack{+3.88\\{\footnotesize [0.88, 6.88]}} & \shortstack{+4.17\\{\footnotesize [3.15, 5.27]}} & \shortstack{-15.40\\{\footnotesize [-16.31, -14.54]}} & \shortstack{+13.30\\{\footnotesize [8.87, 17.91]}} & \shortstack{+35.31\\{\footnotesize [34.22, 36.44]}}\\
No echo reward & \shortstack{+30.98\\{\footnotesize [29.73, 32.23]}} & \shortstack{-3.50\\{\footnotesize [-4.21, -2.78]}} & \shortstack{+18.14\\{\footnotesize [17.44, 18.85]}} & \shortstack{-0.71\\{\footnotesize [-2.48, 0.89]}} & \shortstack{+12.88\\{\footnotesize [11.11, 14.66]}}\\
No coverage reward & \shortstack{+8.43\\{\footnotesize [7.00, 9.85]}} & \shortstack{-35.78\\{\footnotesize [-36.38, -35.14]}} & \shortstack{-10.30\\{\footnotesize [-11.05, -9.54]}} & \shortstack{+7.62\\{\footnotesize [4.43, 11.17]}} & \shortstack{-34.72\\{\footnotesize [-35.83, -33.58]}}\\
Verdict reward only & \shortstack{+15.68\\{\footnotesize [13.82, 17.57]}} & \shortstack{-31.11\\{\footnotesize [-31.68, -30.53]}} & \shortstack{-19.34\\{\footnotesize [-19.83, -18.88]}} & \shortstack{+0.71\\{\footnotesize [-1.24, 2.66]}} & \shortstack{-32.42\\{\footnotesize [-33.79, -31.05]}}\\
\bottomrule
\end{tabular}
\caption{PAPT trades repeated acquisition against coverage and transfer accuracy. Entries show paired differences in percentage points with 95\% claim-bootstrap intervals. Synthetic comparisons use 200 claims and C1/C2/C3/C2R: five seeds for M1 and reward ablations, common seed zero for M2/M3. SciFact uses 188 claims, three one-document conditions and seed zero.}
\label{tab:pairedpapt}
\end{table*}

\begin{figure*}[t]
\centering
\includegraphics[width=\textwidth]{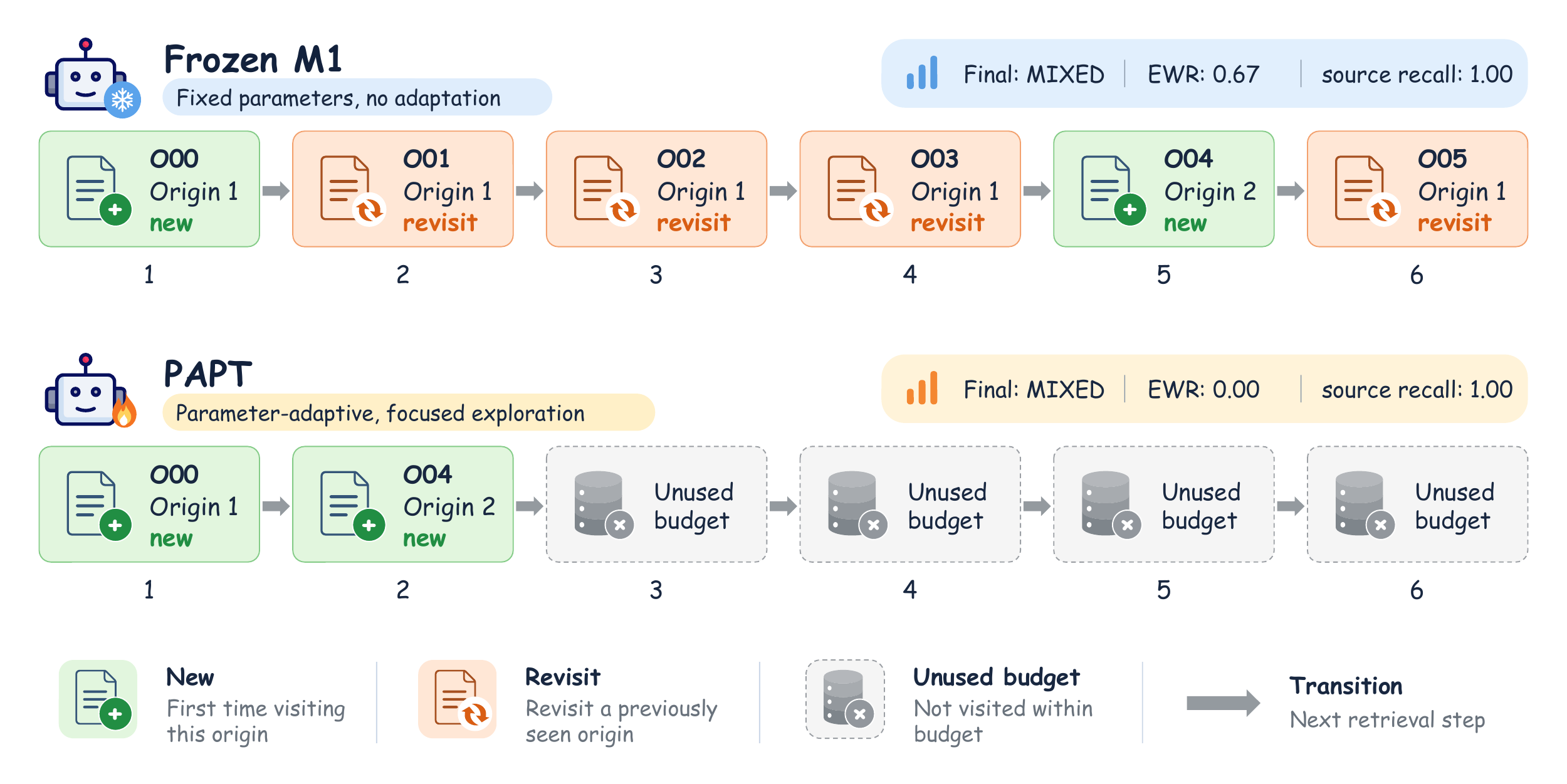}
\caption{PAPT stops after two distinct origins in this C2 example. The matched seed-zero pair lies nearest the median EWR change; both policies return the correct MIXED verdict.}
\label{fig:trajectory}
\end{figure*}

PAPT improves synthetic agent accuracy over frozen Qwen3-4B by 41.5 percentage points across five presentation seeds (Table~\ref{tab:pairedpapt}). It also reduces repeated acquisition, cutting echo walk rate by 36.9 points while using about three fewer walks. Table~\ref{tab:agentic} reports the inference controls on their common seed zero. In the balanced conflict in Figure~\ref{fig:trajectory}, PAPT visits each origin once and stops.

The shorter trajectories also leave more evidence unseen. C3 drives the recall loss: PAPT reaches about one-third of the six available origins, whereas the frozen policy reaches more than four-fifths. PAPT averages roughly two walks, leaving most origins unvisited even though each origin is reachable in one action. These results link reduced repetition to earlier stopping, which also leaves evidence unseen.

Provenance collapse provides a strong cheaper control. On common seed zero, PAPT exceeds M3 accuracy by about four percentage points, but revisits more evidence and reaches fewer distinct origins. M3 changes the action space by grouping equivalent evidence, so this comparison assesses the complete preprocessing intervention. The static results show a further distinction: collapse nearly eliminates redundant-path sensitivity while retaining a response to distinct sources. PAPT still responds more to redundant paths than to additional origins. Appendix~\ref{app:mitigation} reports the full static mitigation and quality estimates.

\subsection{Transfer reduces echoes but loses accuracy}
SciFact exposes a quality cost that the synthetic aggregate conceals. PAPT reduces echo walk rate by 36.0 percentage points relative to frozen Qwen3-4B, yet loses 3.6 points in accuracy (Table~\ref{tab:pairedpapt}). Against provenance collapse, PAPT attains higher accuracy but repeats more evidence. Thus neither intervention dominates both outcomes, and fewer revisits alone provide an incomplete measure of transfer.

All three common SciFact agent conditions contain one document origin. Every compared Qwen3-4B policy reaches that origin, placing source recall at its ceiling. This test measures repeated acquisition and final judgment on scientific rationales; it provides no multi-source acquisition test. The static document-count analysis uses ten eligible multi-document claims and yields correspondingly wide intervals (Appendix~\ref{app:mitigation}).

At this recall ceiling, fewer revisits change repeated exposure without expanding the acquired origins. The accuracy loss therefore calls for evaluating how the policy uses scientific evidence.

\newpage
\subsection{Reward ablations expose competing effects}
The full reward improves synthetic accuracy over verdict-only training by about 16 percentage points, while reducing both echo rate and source recall (Table~\ref{tab:pairedpapt}). Removing the echo term also lowers accuracy relative to full PAPT. Removing coverage yields higher recall than full PAPT, despite receiving no coverage reward. These outcomes show why reward components require joint evaluation: changing one term alters the resulting acquisition policy across several measures.

On SciFact, the full reward offers no clear accuracy advantage over verdict-only training: the paired interval includes zero. The evidence thus supports a synthetic accuracy benefit, while leaving that benefit uncertain in transfer. All variants share a training seed and development selection objective; these comparisons characterize the resulting policies under that shared training seed.

PAPT retains positive accuracy differences over the frozen agent under budgets 3, 5, and 8, ring topology, and variable path lengths (Appendix~\ref{app:sensitivity}). Echo rate falls in all five settings. Source recall increases slightly at budget three and falls in the remaining settings. The acquisition tradeoff therefore extends beyond the primary topology.

\subsection{Stopping depends on unobserved sources}
The C2 and C3 constructions expose a stopping ambiguity. After acquiring one supporting and one refuting origin, an agent has observed balanced evidence in either condition. That pair completes C2's two-origin graph, whereas C3 still contains four unseen supporting origins. The full-graph verdict is therefore MIXED in C2 and SUPPORT in C3. A stopping rule based only on the balance of acquired evidence would treat these states alike. Source recall identifies the difference because its denominator includes the distinct origins that remain available. This comparison explains why correct source counting over observed evidence can coexist with an incorrect final judgment.

Provenance collapse and policy training intervene at different points in this decision. Collapse removes equivalent choices before exploration, while PAPT selects among the original graph's neighbors and can end its trajectory early. Their accuracy differences can therefore reflect both evidence selection and stopping behavior. A follow-up comparison could give both policies the same set of distinct origins before eliciting their verdicts. Matching the acquired evidence would separate differences in judgment from differences in evidence access. Complementary tests could vary the stopping budget on a fixed graph to examine which origins each policy prioritizes.

\section{Conclusion and Future Work}
\graphecho{} separated structural repetition from provenance through matched graph interventions. Frozen agents showed heterogeneous judgment effects and consistently higher echo walk rates under redundant support. PAPT improved synthetic accuracy and reduced repeated walks, but lowered source recall and SciFact accuracy. Evaluation must measure both repeated and unobserved evidence. Future work should test multi-source acquisition on scientific graphs across training seeds.

Future scientific benchmarks could track shared datasets and experiments across papers. These annotations would help test whether agents recognize common empirical origins across documents. They would also support comparisons between evidence grouping and learned exploration when document boundaries provide an incomplete account of source dependence.

\newpage
\section*{Limitations of the Work}
Synthetic source independence is a controlled benchmark assumption. Equal-weight evidence balance defines the synthetic verdict labels, while real scientific evidence varies in strength and dependence. Repeating identical evidence text controls information content but covers a limited form of graph extraction. SciFact's document identifiers provide an observable provenance proxy, and its gold-rationale setting omits retrieval errors. Self-reported probabilities measure model responses without establishing probabilistic calibration. The agentic results concern the implemented graph topologies, previews, and action costs. Shorter trajectories can improve echo rate while leaving distinct evidence unobserved.

Training uses one random seed. Claim-bootstrap intervals quantify evaluation variation conditional on the fitted policy and omit variation across training runs. The reward ablations therefore compare the resulting policies under a shared training seed.

\section*{Declaration on Generative AI}
The authors used generative AI to polish the language of the manuscript. The authors retain full responsibility for its content.

\clearpage
\bibliography{references}
\clearpage
\appendix
\section{Data construction checks}
\graphecho{}-Syn keeps claims and entity names disjoint across training, development, and test splits. Construction checks verify atom counts, path counts, path lengths, and matched evidence content across interventions. The SciFact conversion preserves the original document identifiers and rationale sets, grouping each document's rationale sentences into one evidential origin. Additional representations retain the atom's source identity and preserve the evidence-balance label.

\section{Static and conflict responses}
\label{app:static}

\begin{figure*}[t]
\centering\includegraphics[width=\textwidth]{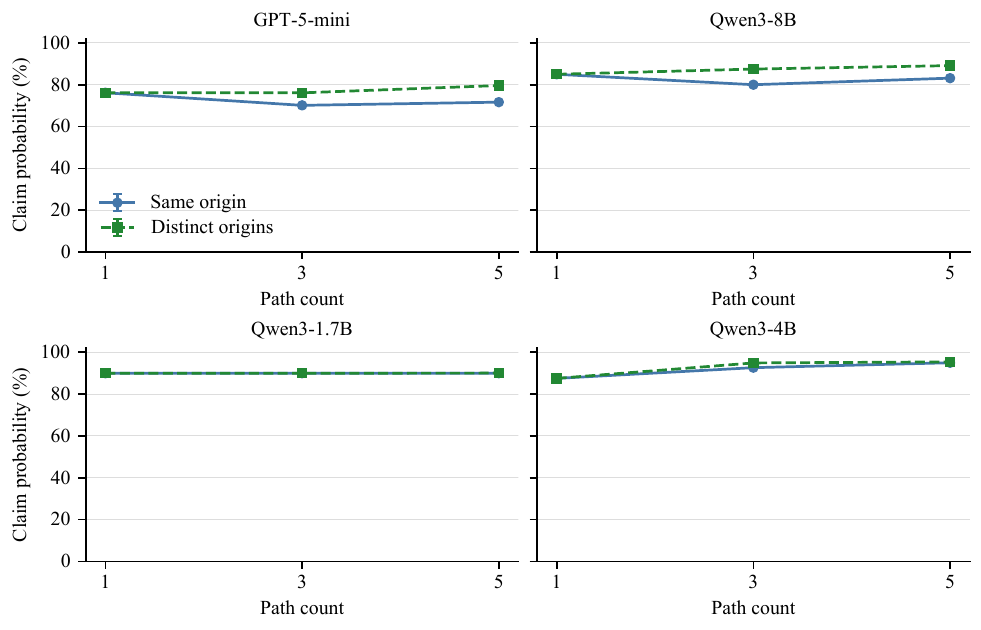}
\caption{Probability responses separate redundant paths from distinct origins. Same-origin curves keep one source; distinct-origin curves match source count to path count. Error bars show 95\% claim-bootstrap intervals over 200 claims and five seeds.}
\label{fig:staticcurves}
\end{figure*}

Table~\ref{tab:conflict} reports the paired conflict contrasts. Figure~\ref{fig:staticcurves} shows all one-sided supporting conditions. The source-count intervention preserves path templates and evidence wording. The comparison therefore attributes response differences to the provenance assignment within this construction, while retaining model-specific directions of change. An unchanged SUPPORT rate can coexist with a probability shift toward redundant evidence.

\section{Static mitigation and task quality}
\label{app:mitigation}

Tables~\ref{tab:mitigation} and~\ref{tab:mitigationquality} use common presentation seed zero for every method. The synthetic setting contains 200 claims and ten conditions. SciFact contains 188 claims and 584 instances; only ten claims contribute to document-count effects. We average each claim's available conditions before pooling task-quality estimates. Document-count effects compare the available two-to-five-document set against one document at a matched path count. A stratum containing one claim provides a descriptive observation without an inferential interval. Within-claim averaging gives each claim equal weight despite different numbers of available variants.

\section{Agent validity and condition-level behavior}
\label{app:diagnostics}

\begin{figure*}[t]
\centering\includegraphics[width=\textwidth]{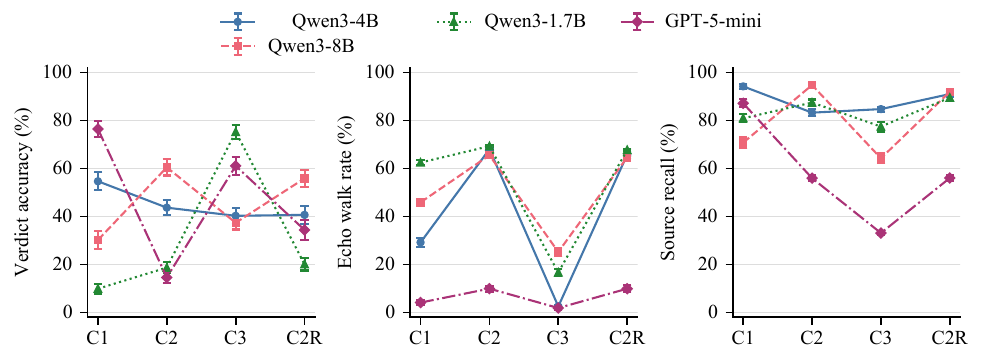}
\caption{Echo-rich conditions affect acquisition and accuracy differently. Frozen agents use 200 claims, five seeds and budget six; error bars show 95\% claim-bootstrap intervals.}
\label{fig:agentconditions}
\end{figure*}

Table~\ref{tab:actiondiagnostics} reports exact operational counts; all five policies attempt at least one walk in every received output. Table~\ref{tab:validity} conditions on final-output validity and therefore describes a selected subset. Figure~\ref{fig:agentconditions} retains all received outputs for accuracy and shows each conflict condition separately. Malformed responses count as incorrect judgments in the primary evaluation. Invalid actions enter the walk denominator; only valid repeated evidence visits enter the echo numerator.

\section{Training and sensitivity settings}
\label{app:sensitivity}

All four policies use the same 500 balanced training instances, one epoch, seed 42, learning rate $10^{-5}$, group size four, LoRA rank 16 and alpha 32. Training samples actions with temperature 0.7 and top-$p$ 0.9. Training uses a zero Kullback--Leibler (KL) coefficient and gradient clipping at norm one. Checkpoint selection compares the policies at training steps 450, 475, and 500. Each candidate receives the same 400-instance development evaluation on C1/C2/C3/C2R. Development full-objective reward selects the test policy. This shared selection criterion evaluates every reward ablation against the same acquisition objective.

\begin{figure*}[t]
\centering\includegraphics[width=\textwidth]{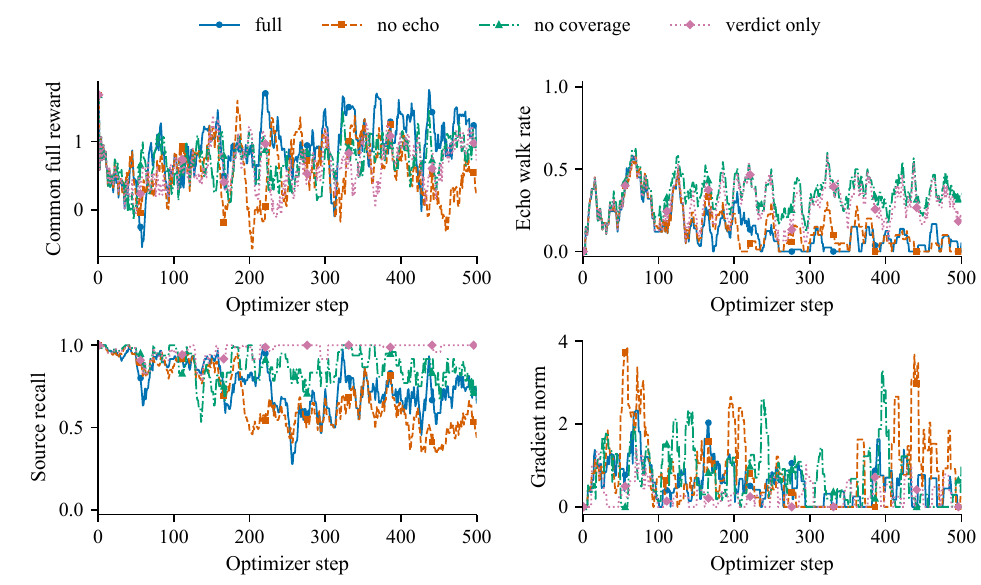}
\caption{Reward variants produce different acquisition behavior during training. Curves use trailing ten-step averages and the full reward for comparison across objectives; the traces describe one training seed.}
\label{fig:training}
\end{figure*}

Figure~\ref{fig:training} shows that full-reward and verdict-only training attain different echo-rate and source-recall trajectories. The curves describe optimization conditional on one training seed.

Table~\ref{tab:sensitivity} pairs the frozen and trained agents under alternate budgets and graph constructions. Ring topology changes adjacency, and the variable-hop construction cycles through two to six hops while preserving evidence and gold labels. All three frozen Qwen models and full PAPT also complete static sampling and variable-hop evaluations. These tests keep training fixed and evaluate sensitivity conditional on the fitted policies.

\section{Supplementary SciFact baselines}

Table~\ref{tab:scifactbaselines} compares the frozen models on scientific rationales and reports the supplementary GPT-5-mini controls. The static and agent columns measure different outcomes and use the stated presentation-seed scopes. The agent columns use 1S5P, where all routes share one document origin. Positive static PME indicates movement toward the gold evidence stance under repetition. Since this agent condition contains one origin, its echo rate describes repeated exposure within a document.

\begin{table}[!h]
\centering\small\setlength{\tabcolsep}{2pt}
\begin{tabular}{lrr}
\toprule
Model & \shortstack{$\Delta$ SUPPORT\\rate} & \shortstack{$\Delta$ claim\\probability}\\
\midrule
Qwen3-1.7B & \shortstack{69.40\\{\footnotesize [66.00, 72.70]}} & \shortstack{27.77\\{\footnotesize [26.41, 29.09]}}\\
Qwen3-4B & \shortstack{0.10\\{\footnotesize [0.00, 0.30]}} & \shortstack{0.10\\{\footnotesize [-0.38, 0.60]}}\\
Qwen3-8B & \shortstack{0.00\\{\footnotesize [0.00, 0.00]}} & \shortstack{4.70\\{\footnotesize [4.01, 5.39]}}\\
GPT-5-mini & \shortstack{0.00\\{\footnotesize [0.00, 0.00]}} & \shortstack{5.46\\{\footnotesize [5.09, 5.84]}}\\
\bottomrule
\end{tabular}
\caption{Conflict effects for C2 minus C1 in percentage points, with 95\% paired base-claim bootstrap intervals.}
\label{tab:conflict}
\end{table}

\begin{table*}[t]
\centering\small\setlength{\tabcolsep}{2pt}
\begin{tabular}{lrrr}
\toprule
Method & Syn. PME & Syn. ISE & SciFact PME$^\dagger$\\
\midrule
M0: source nodes & \shortstack{7.08\\{\footnotesize [6.72, 7.42]}} & \shortstack{0.53\\{\footnotesize [0.33, 0.75]}} & \textit{n/a}\\
M1: provenance & \shortstack{7.47\\{\footnotesize [7.12, 7.83]}} & \shortstack{0.30\\{\footnotesize [0.15, 0.47]}} & \shortstack{1.91\\{\footnotesize [-0.69, 4.60]}}\\
M2: instruction & \shortstack{7.22\\{\footnotesize [6.88, 7.58]}} & \shortstack{1.95\\{\footnotesize [1.62, 2.30]}} & \shortstack{1.41\\{\footnotesize [-1.25, 3.99]}}\\
M3: collapse & \shortstack{0.05\\{\footnotesize [-0.07, 0.20]}} & \shortstack{5.90\\{\footnotesize [5.65, 6.17]}} & \shortstack{0.13\\{\footnotesize [-0.56, 0.98]}}\\
PAPT & \shortstack{7.35\\{\footnotesize [7.00, 7.70]}} & \shortstack{0.28\\{\footnotesize [0.12, 0.45]}} & \shortstack{1.06\\{\footnotesize [-1.62, 3.83]}}\\
No echo reward & \shortstack{5.80\\{\footnotesize [5.55, 6.05]}} & \shortstack{0.68\\{\footnotesize [0.45, 0.93]}} & \shortstack{1.46\\{\footnotesize [-1.06, 4.07]}}\\
No coverage reward & \shortstack{8.18\\{\footnotesize [7.75, 8.60]}} & \shortstack{2.60\\{\footnotesize [2.27, 2.95]}} & \shortstack{0.03\\{\footnotesize [-2.93, 3.11]}}\\
Verdict reward only & \shortstack{6.10\\{\footnotesize [5.83, 6.40]}} & \shortstack{2.62\\{\footnotesize [2.27, 2.98]}} & \shortstack{1.60\\{\footnotesize [-1.54, 4.81]}}\\
\bottomrule
\end{tabular}
\caption{Qwen3-4B static mitigation and transfer effects in percentage points. All methods use presentation seed zero. The dagger denotes stance alignment. Brackets show 95\% base-claim bootstrap intervals; n/a denotes an unscheduled control.}
\label{tab:mitigation}
\end{table*}

\begin{table*}[t]
\centering\small\setlength{\tabcolsep}{2pt}
\begin{tabular}{lrrrrr}
\toprule
Method & Syn. accuracy & Syn. count error & SciFact accuracy & SciFact count error & SciFact DPE\\
\midrule
M0: source nodes & \shortstack{82.70\\{\footnotesize [81.90, 83.50]}} & \shortstack{1.17\\{\footnotesize [1.13, 1.21]}} & \textit{n/a} & \textit{n/a} & \textit{n/a}\\
M1: provenance & \shortstack{81.25\\{\footnotesize [80.60, 81.90]}} & \shortstack{1.18\\{\footnotesize [1.13, 1.22]}} & \shortstack{75.78\\{\footnotesize [70.04, 81.13]}} & \shortstack{0.94\\{\footnotesize [0.86, 1.02]}} & \shortstack{2.50\\{\footnotesize [-3.50, 10.50]}}\\
M2: instruction & \shortstack{79.40\\{\footnotesize [78.50, 80.30]}} & \shortstack{1.15\\{\footnotesize [1.10, 1.20]}} & \shortstack{74.75\\{\footnotesize [68.94, 80.35]}} & \shortstack{0.87\\{\footnotesize [0.80, 0.94]}} & \shortstack{4.50\\{\footnotesize [-0.50, 12.00]}}\\
M3: collapse & \shortstack{80.45\\{\footnotesize [79.60, 81.25]}} & \shortstack{0.85\\{\footnotesize [0.81, 0.88]}} & \shortstack{74.18\\{\footnotesize [67.80, 80.04]}} & \shortstack{0.65\\{\footnotesize [0.56, 0.74]}} & \shortstack{11.00\\{\footnotesize [0.00, 23.50]}}\\
PAPT & \shortstack{81.60\\{\footnotesize [81.05, 82.20]}} & \shortstack{0.99\\{\footnotesize [0.96, 1.03]}} & \shortstack{74.08\\{\footnotesize [68.16, 79.65]}} & \shortstack{0.89\\{\footnotesize [0.84, 0.94]}} & \shortstack{4.50\\{\footnotesize [-0.50, 12.00]}}\\
No echo reward & \shortstack{80.35\\{\footnotesize [79.45, 81.25]}} & \shortstack{1.11\\{\footnotesize [1.09, 1.14]}} & \shortstack{76.42\\{\footnotesize [70.74, 81.70]}} & \shortstack{0.86\\{\footnotesize [0.79, 0.92]}} & \shortstack{2.00\\{\footnotesize [-3.50, 10.00]}}\\
No coverage reward & \shortstack{81.10\\{\footnotesize [80.60, 81.55]}} & \shortstack{0.90\\{\footnotesize [0.86, 0.94]}} & \shortstack{67.06\\{\footnotesize [60.89, 73.12]}} & \shortstack{0.76\\{\footnotesize [0.67, 0.86]}} & \shortstack{3.00\\{\footnotesize [-1.00, 9.50]}}\\
Verdict reward only & \shortstack{81.20\\{\footnotesize [80.70, 81.75]}} & \shortstack{1.10\\{\footnotesize [1.06, 1.14]}} & \shortstack{70.89\\{\footnotesize [64.86, 76.67]}} & \shortstack{0.86\\{\footnotesize [0.78, 0.94]}} & \shortstack{4.50\\{\footnotesize [-0.50, 12.00]}}\\
\bottomrule
\end{tabular}
\caption{Task quality accompanies redundancy sensitivity for Qwen3-4B. All methods use presentation seed zero. Accuracy uses percent. DPE compares each available document set against one document at matched path counts, pooling the eligible two-to-five-document claims. Brackets show 95\% base-claim bootstrap intervals.}
\label{tab:mitigationquality}
\end{table*}

\begin{table*}[t]
\centering\small\setlength{\tabcolsep}{3pt}
\begin{tabular}{lrrrr}
\toprule
Policy & Connector walks & Invalid attempts & Zero-walk outputs & Valid finals\\
\midrule
Qwen3-1.7B & 404 & 447 & 0 & 3059\\
Qwen3-4B & 0 & 524 & 0 & 3992\\
Qwen3-8B & 0 & 2151 & 0 & 3961\\
GPT-5-mini & 0 & 27 & 0 & 3999\\
PAPT & 0 & 0 & 0 & 4000\\
\bottomrule
\end{tabular}
\caption{Operational counts over 4,000 synthetic agent outputs per policy (200 claims, four conditions and five seeds). Counts describe received outputs; a valid output format does not guarantee a correct verdict.}
\label{tab:actiondiagnostics}
\end{table*}

\begin{table*}[t]
\centering\small
\begin{tabular}{lrrr}
\toprule
Model & Valid finals C1/C2 & Both-valid claims & Both-valid accuracy change (pp)\\
\midrule
Qwen3-1.7B & 391/905 & 168 & $-2.88$ [$-10.59$, 4.73]\\
Qwen3-4B & 996/998 & 200 & $-10.97$ [$-15.78$, $-6.18$]\\
Qwen3-8B & 961/1000 & 200 & 28.43 [23.38, 33.55]\\
GPT-5-mini & 1000/999 & 200 & $-61.90$ [$-65.80$, $-58.00$]\\
\bottomrule
\end{tabular}
\caption{Final-output validity changes the population available for a diagnostic. Each condition contains 1,000 outputs. The last column gives C2-minus-C1 accuracy among matched pairs with valid finals on both sides, with 95\% claim-bootstrap intervals.}
\label{tab:validity}
\end{table*}

\begin{table*}[t]
\centering\small\setlength{\tabcolsep}{3pt}
\begin{tabular}{lrrr}
\toprule
Evaluation variant & $\Delta$ accuracy & $\Delta$ echo rate & $\Delta$ source recall\\
\midrule
Budget 3 & \shortstack{+18.25\\{\footnotesize [14.88, 21.62]}} & \shortstack{-23.15\\{\footnotesize [-24.40, -21.94]}} & \shortstack{+1.33\\{\footnotesize [0.06, 2.60]}}\\
Budget 5 & \shortstack{+39.62\\{\footnotesize [35.38, 43.62]}} & \shortstack{-35.91\\{\footnotesize [-37.11, -34.67]}} & \shortstack{-6.88\\{\footnotesize [-8.40, -5.38]}}\\
Budget 8 & \shortstack{+23.50\\{\footnotesize [19.75, 27.25]}} & \shortstack{-33.73\\{\footnotesize [-34.92, -32.57]}} & \shortstack{-15.81\\{\footnotesize [-16.96, -14.67]}}\\
Ring topology & \shortstack{+21.25\\{\footnotesize [17.00, 25.50]}} & \shortstack{-28.87\\{\footnotesize [-30.57, -27.15]}} & \shortstack{-17.17\\{\footnotesize [-18.94, -15.40]}}\\
Variable hops & \shortstack{+34.62\\{\footnotesize [30.00, 39.25]}} & \shortstack{-32.68\\{\footnotesize [-34.01, -31.34]}} & \shortstack{-11.08\\{\footnotesize [-12.58, -9.56]}}\\
\bottomrule
\end{tabular}
\caption{PAPT-minus-frozen agent differences across acquisition settings. Each comparison uses 200 claims, C1/C2/C3/C2R and seed zero. Entries show percentage points and 95\% paired claim-bootstrap intervals. Ring and variable-hop evaluations use budget six.}
\label{tab:sensitivity}
\end{table*}

\begin{table*}[t]
\centering\small\setlength{\tabcolsep}{3pt}
\begin{tabular}{lrrr}
\toprule
Model / control & Static PME (pp) & Agent accuracy (\%) & Agent echo rate (\%)\\
\midrule
Qwen3-1.7B & \shortstack{+0.98\\{\footnotesize [-0.59, 2.69]}} & \shortstack{40.43\\{\footnotesize [33.51, 47.34]}} & \shortstack{77.07\\{\footnotesize [75.98, 78.10]}}\\
Qwen3-4B & \shortstack{+1.73\\{\footnotesize [-0.72, 4.23]}} & \shortstack{78.19\\{\footnotesize [72.34, 84.04]}} & \shortstack{79.92\\{\footnotesize [79.81, 80.00]}}\\
Qwen3-8B & \shortstack{-3.22\\{\footnotesize [-6.42, -0.19]}} & \shortstack{80.85\\{\footnotesize [75.00, 86.17]}} & \shortstack{79.20\\{\footnotesize [78.49, 79.84]}}\\
GPT-5-mini: M1 & \shortstack{-0.46\\{\footnotesize [-1.29, 0.30]}} & \shortstack{84.57\\{\footnotesize [79.26, 89.36]}} & \shortstack{16.49\\{\footnotesize [13.03, 20.04]}}\\
GPT-5-mini: M2 & \shortstack{-0.04\\{\footnotesize [-1.18, 1.05]}} & \shortstack{85.11\\{\footnotesize [79.79, 89.89]}} & \shortstack{16.84\\{\footnotesize [13.56, 20.30]}}\\
GPT-5-mini: M3 & \shortstack{-0.65\\{\footnotesize [-1.44, 0.14]}} & \shortstack{88.30\\{\footnotesize [83.51, 92.55]}} & \shortstack{0.00\\{\footnotesize [0.00, 0.00]}}\\
\bottomrule
\end{tabular}
\caption{SciFact frozen baselines and supplementary API controls. Static PME aligns the redundant-path effect with gold evidence stance; M1 uses five presentation seeds and M2/M3 use seed zero. Agent columns show the 1S5P condition at seed zero. Agent estimates use 188 claims. Static PME uses 188 claims except GPT-5-mini M3, which retains 187 claims with valid paired outputs. Brackets show 95\% claim-bootstrap intervals.}
\label{tab:scifactbaselines}
\end{table*}

\end{document}